\documentclass[]{amap}

\usepackage[toc,page,header]{appendix}
\usepackage{minitoc}
\usepackage{solarized-light}
\usepackage{booktabs}
\usepackage{multirow}
\usepackage{graphicx}
\usepackage{array}
\usepackage{siunitx,array}
\usepackage[table]{xcolor}
\usepackage{makecell,array,booktabs}
\usepackage{makecell}
\usepackage{framed}

\usepackage{bbding}
\usepackage{hyperref}
\usepackage{amsmath}
\usepackage{amsfonts}
\usepackage{placeins}
\usepackage[colorinlistoftodos]{todonotes}
\usepackage{longtable}
\usepackage{hhline}
\usepackage{fancyvrb}
\usepackage{graphicx}
\usepackage[table]{xcolor}
\usepackage{booktabs}
\usepackage{float}
\usepackage{fvextra}
\usepackage{CJKutf8}
\usepackage{multicol}
\usepackage{float}
\usepackage{placeins}
\usepackage{cleveref}
\usepackage{tablefootnote}
\usepackage{threeparttable}
\usepackage{tabularx}
\usepackage{mdframed}
\usepackage{subcaption}
\usepackage[usestackEOL]{stackengine}
\usepackage[numbers]{natbib}
\newcommand{\commentout}[1]{}
\renewcommand{\paragraph}[1]{\noindent\textbf{#1.}\hspace*{1em}}
\usepackage{enumitem}
\usepackage{hyperref}
\setlist[itemize]{leftmargin=15pt}
\usepackage{siunitx,array}
\usepackage{tabularx}
\usepackage{adjustbox}
\usepackage{arydshln}
\usepackage{pifont}
\usepackage{bbding}
\definecolor{ampblue}{rgb}{0.82, 0.88, 0.94}
\usepackage[table]{xcolor} 
\usepackage{array} 
\usepackage[dvipsnames]{xcolor}
\usepackage{fontawesome5}
\usepackage{ragged2e}   
\newcolumntype{Y}{>{\RaggedRight\arraybackslash}X}

\RequirePackage{xspace}
\makeatletter
\DeclareRobustCommand\onedot{\futurelet\@let@token\@onedot}
\def\@onedot{\ifx\@let@token.\else.\null\fi\xspace}

\makeatother

\usepackage{xcolor}

\newcolumntype{C}{>{\centering\arraybackslash}p{1.2cm}}
\newcolumntype{D}{>{\centering\arraybackslash}p{2.0cm}}

\title{ABot-OCR Technical Report}

\author{AMAP CV Lab}
\vspace{-11pt}

\contribution{Kaitao Jiang, Ruiyan Gong, Xiaolong Cheng, Kangning Niu, Tianlun Li, Mu Xu.}

\abstract{

We introduce ABot-OCR, an end-to-end vision-language model that transcribes a page image directly into clean Markdown in a single forward pass. By doing so, our approach completely eliminates the need for brittle modular orchestration.
To maximize parsing fidelity, we develop a dedicated data engine to provide large-scale, structurally consistent supervision. Furthermore, we propose Decoupled Heterogeneous Document Optimization, a structure-constrained reinforcement learning method that sharpens textual accuracy and strictly enforces markup well-formedness beyond supervised fine-tuning alone.
Extensive evaluations demonstrate the superior performance of our framework. On the OmniDocBench v1.5 and v1.6 benchmarks, ABot-OCR achieves state-of-the-art scores of 92.81 and 93.30 among all end-to-end systems, substantially narrowing the performance gap relative to strong pipeline baselines. Finally, comprehensive multilingual text recognition across ten diverse languages further confirms the robust generalizability of ABot-OCR.

\bigskip

\textbf{Code:} \url{https://github.com/amap-cvlab/ABot-OCR}

\textbf{Model:} \url{https://huggingface.co/acvlab/ABot-OCR}

}

\begin{document}
\maketitle
\vspace{-4pt}

\vspace{0.5em}
\noindent
\begin{center}
  \includegraphics[width=0.88\textwidth]{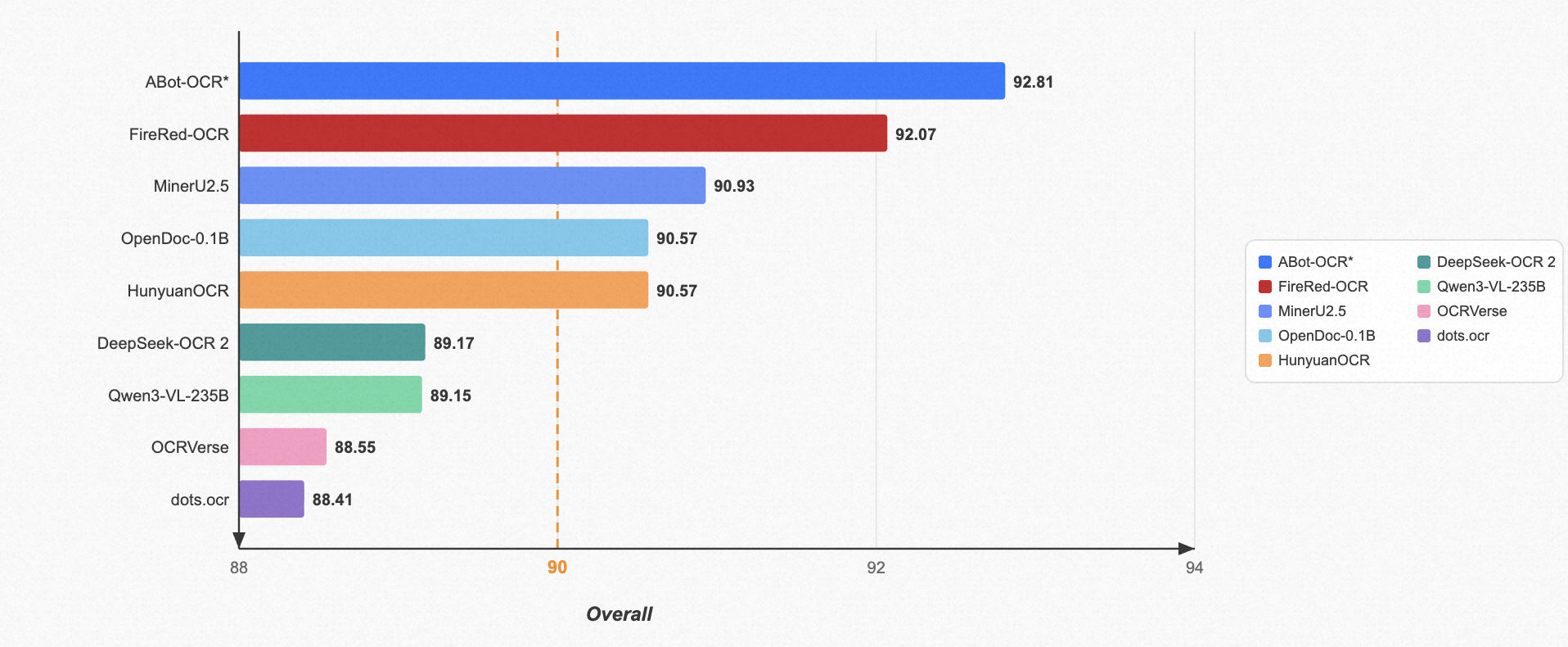}
  \captionof{figure}{Performance comparison of document parsing methods on OmniDocBench v1.5 across overall.}
  \label{fig:omnidocbench1.5-comparison}
\end{center}
\vspace{0.8em}




\providecommand{\etal}{et al.}
\providecommand{\argmax}{\operatornamewithlimits{arg\,max}}
\providecommand{\Lev}{\operatorname{Lev}}
\providecommand{\KL}{\operatorname{KL}}
\providecommand{\clip}{\operatorname{clip}}
\providecommand{\Rtext}{R_{\text{text}}}
\providecommand{\Rformula}{R_{\text{formula}}}
\providecommand{\Rtable}{R_{\text{table}}}
\providecommand{\Rstruct}{R_{\text{struct}}}

\section{Introduction}

Document parsing, a fundamental task in computer vision, aims to transform visually unstructured content into textually structured formats (e.g., Markdown) to facilitate downstream applications such as information extraction. However, this task remains an open challenge, primarily due to the high variability of document layouts and the complex structural compositions of elements like tables and formulas. Conventionally, pipeline-based methods \cite{cui2025paddleocr30technicalreport, livathinos2025docling, zhao2024doclayout} decompose document parsing into a sequence of distinct subtasks. Following the advent of GPT-4v \cite{openai2023gpt}, the remarkable generalization capabilities of Vision-Language Models (VLMs) have garnered significant attention, inspiring researchers to pioneer the exploration of VLM-based document parsing. Currently, VLM-based document parsing diverges into two primary paradigms. The first is the Decoupled Paradigm \cite{feng2025dolphin, niu2025mineru2}, which decouples layout analysis from content recognition by treating them as distinct subtasks of VLMs. In contrast, the second is the End-to-End Paradigm \cite{wei2024general, hunyuanocr, wang2026mineru2}, which directly converts the document image into a structured format in a single step. Overall, traditional pipeline-based models continue to maintain their predominance in document parsing. However, given the unprecedented pace of advancements in Vision-Language Models (VLMs) \cite{qwen3.5, hong2025glm, guo2025seed1}, we envision the End-to-End paradigm, underpinned by the exponentially growing capabilities of VLMs, as the ultimate form of document parsing, poised to become the mainstream approach in the near future.

We present \textbf{Abot-OCR}, a 2B-parameter vision-language model for document parsing. By achieving the-state-of-art performance on both OmniDocBench v1.5 \cite{omnidocbench2024} and OmniDocBench v1.6 \cite{mineru25pro} among VLM-based approaches, Abot-OCR bridges the gap between end-to-end models and pipeline-based methods. Additionally, we also extend our exploration to multilingual OCR, evaluating the performance on document images across 10 additional languages. These include four UN official languages (Arabic, Spanish, French and Russian) and six additional languages (German, Japan, Korean, Portug, Thai and Vietna).



The effectiveness of VLM-based document parsing models depends not only on data scale, but also on the quality, consistency, and verifiability of annotations. Therefore, we further build a dedicated data engine for Abot-OCR, consisting of \textbf{Hierarchical-Consistency Annotation Verification} and \textbf{Web-Scale Document Pseudo-Labeling}. For existing annotated data, the verification process follows a cost-aware coarse-to-fine hierarchy: lightweight linguistic consistency checks are first applied to filter malformed annotations without accessing document images; samples that pass this stage are then examined by visual consistency verification through layout analysis and expert recognizers; finally, since the preceding linguistic and visual checks rely on predefined rules and matching strategies. Therefore, we further introduce a VLM reasoner as the final verification stage to assess whether the annotated document content and document structures are semantically and visually consistent with the input image. For unlabeled web documents, we adopt a modular pseudo-labeling pipeline that decomposes pages by layout analysis, labels regions with task-specific expert models, and assembles structured page-level pseudo-labels. Both verified annotations and generated pseudo-labels are unified by DPCS-based quality control, improving training data coverage while preserving high label quality.

In terms of methodology, we propose  a progressive three-stage training strategy for Abot-OCR. Stage 1 builds modular document parsing capabilities, including text spotting, formula recognition, table recognition, and layout analysis, thereby strengthening both fine-grained perception and basic structural awareness. Stage 2 unifies these abilities through end-to-end page-level parsing in the format of Markdown. Stage 3 introduces a novel document structure-constrained reinforcement learning framework: \textbf{Decoupled Heterogeneous Document Optimization (DHDO)}. DHDO introduces four distinct verifiable rewards, including a perception rewards and three structure rewards. To improve structural capability without degrading recognition accuracy, structure rewards are activated only when the perception reward exceeds a reliable threshold. DHDO further avoids advantage collapse in naive GRPO-style multi-reward normalization by independently normalizing each reward component before aggregation, followed by batch-level rescaling. This preserves fine-grained reward signals and enables more faithful structural optimization. Experimental results show that DHDO consistently outperforms conventional GRPO-style optimization on document parsing tasks, demonstrating the effectiveness of perception-conditioned rewards and decoupled reward normalization for structure-aware OCR training.

\section{Data Engine}
\label{sec:data-engine}

The performance of OCR-oriented multimodal large models is increasingly limited by the quality, coverage, and verifiability of training data; therefore,  we build a novel data engine composed of \textit{Hierarchical-Consistency Annotation Verification} and \textit{Web-Scale Document Pseudo-Labeling}. 

\begin{figure}[htbp] 
  \centering
  \includegraphics[width=0.9\textwidth]{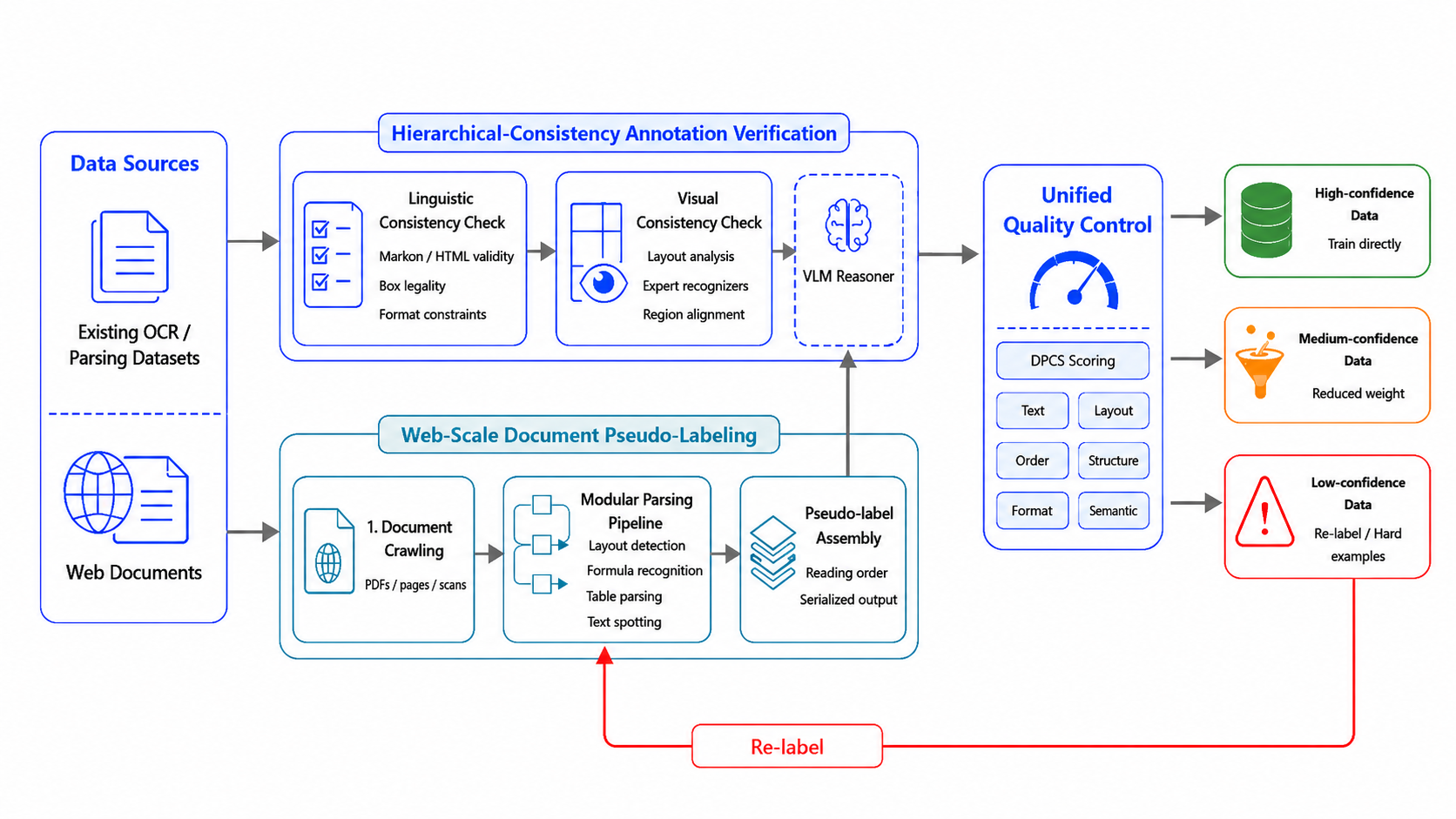}
  \caption{\textbf{Overview of the proposed data engine for Abot-OCR.} The framework combines hierarchical-consistency annotation verification and web-scale document pseudo-labeling to improve the quality and coverage of OCR-oriented multimodal training data. Verified annotations and pseudo-labels are assessed by unified quality control and categorized into different confidence levels, while low-confidence samples are routed back for re-labeling and pipeline refinement.}
  \label{fig:data-engine}
\end{figure}


\subsection{Hierarchical-Consistency Annotation Verification}
\label{sec:annotated-data-verification}

Existing OCR and document parsing datasets are often collected from heterogeneous sources with different annotation formats and quality standards. Directly mixing such data may introduce inconsistent supervision, especially for structured document parsing. We therefore verify each annotated sample from two complementary perspectives: linguistic consistency and visual consistency.

\paragraph{Linguistic consistency verification}
This stage checks whether the annotation is well-formed without referring to the document image. For end-to-end document parsing data, we validate whether the serialized output conforms to the target representation. For instance, \LaTeX\ expressions are checked for balanced delimiters and compilable syntax, while HTML tables are parsed to ensure valid row-column structures and properly nested tags. For text spotting data, we verify bounding-box legality, including non-negative coordinates, valid areas, page-boundary constraints, and abnormal overlaps. 


\paragraph{Visual consistency verification}
This stage checks whether the annotation is consistent with the document image. Since current VLMs are not always reliable as direct end-to-end document parsers, we use a hybrid verifier that combines pipeline-based parsing with VLM-based reasoning. For document parsing samples, a layout analysis model first segments the page into functional regions. Each region is then routed to a task-specific expert recognizer, and the recognized regional outputs are aligned with the original annotation according to location, reading order, and structural type. The verifier computes consistency scores including layout agreement, bounding-box alignment, text similarity, table-structure equivalence, formula-syntax validity, and reading-order consistency.

\paragraph{VLM-assisted consistency scoring} A multimodal reasoning model with large parametric size is used as an auxiliary judge. Given the document image, the candidate annotation, and optionally pipeline-derived intermediate results, the reasoner evaluates whether the annotation is visually and structurally consistent with the image. We summarize this judgment using a Document Parsing Consistency Score (DPCS):
\[
S_{\mathrm{DPCS}} =
25S_{\mathrm{text}} +
15S_{\mathrm{layout}} +
15S_{\mathrm{order}} +
20S_{\mathrm{structure}} +
15S_{\mathrm{format}} +
10S_{\mathrm{semantic}},
\]
where each sub-score is normalized to $[0,1]$. The six terms respectively measure text fidelity, layout localization, reading order, structural fidelity, format validity, and semantic completeness. This dimension-wise design avoids relying on a single binary VLM decision and makes the quality assessment more interpretable.

\paragraph{Unified quality control}
The VLM reasoner produces a Document Parsing Consistency Score (DPCS) that measures the visual and structural consistency between the document image and the candidate annotation. Samples with $S_{\mathrm{DPCS}} \geq 80$ are treated as high-confidence data and retained for training. Samples with scores between 80 and  60 are retained with reduced training weight or task-specific caution. Samples below 60 are considered unreliable and sent for re-labeling before they can be used in training. For subtask datasets, the scoring weights are adjusted according to task-specific reliability requirements.

\subsection{Web-Scale Document Pseudo-Labeling}
\label{sec:web-document-annotation}

Beyond verifying existing annotations, the data engine constructs new training data from large-scale web documents. Since pipeline-based systems remain more stable than fully generative VLMs for many structured OCR tasks, we use a modular pseudo-labeling pipeline as the primary annotation mechanism.

For end-to-end document parsing data, each crawled document is converted into page images and processed by a layout analysis model. Detected regions are categorized into functional types. Each region is annotated by the corresponding expert model, and the regional outputs are assembled into a page-level serialized representation according to layout relationships and predicted reading order.

For sub-task data, we use task-specific annotation pipelines. Formula recognition data are generated by detecting formula regions and applying specialized formula recognizers, followed by \LaTeX\ normalization and syntax checking. Table parsing data are produced by table detection and structure recognition models, followed by canonicalization of cell spans, row-column alignment, and HTML/Markdown normalization. Text spotting data are generated or refined by expert OCR detectors and recognizers, followed by bounding-box legality checks and transcription consistency verification.

All pseudo-labeled samples are passed through DPCS-based gating described in Sec.~\ref{sec:annotated-data-verification}. This shared design ensures that existing annotations and newly generated web labels are evaluated under a unified quality standard. Low-confidence pseudo-labels are filtered rather than directly injected into the training corpus.

\providecommand{\etal}{et al.}
\providecommand{\argmax}{\operatornamewithlimits{arg\,max}}
\providecommand{\Lev}{\operatorname{Lev}}
\providecommand{\KL}{\operatorname{KL}}
\providecommand{\Rtext}{R_{\text{text}}}
\providecommand{\Rformula}{R_{\text{formula}}}
\providecommand{\Rtable}{R_{\text{table}}}
\providecommand{\Rstruct}{R_{\text{struct}}}

\section{Training Strategy}
\label{sec:method}

In the task of document parsing, there are two primary capabilities. The first is \textbf{perception capability}, which refers to the ability to accurately and comprehensively recognize all content in the correct reading order. The second is \textbf{structural capability}, which denotes the capability to faithfully reconstruct the document layout, ensuring that the correct content is placed within the appropriate structural context.

\begin{figure}[htbp]
    \centering
    \includegraphics[width=1.0\linewidth]{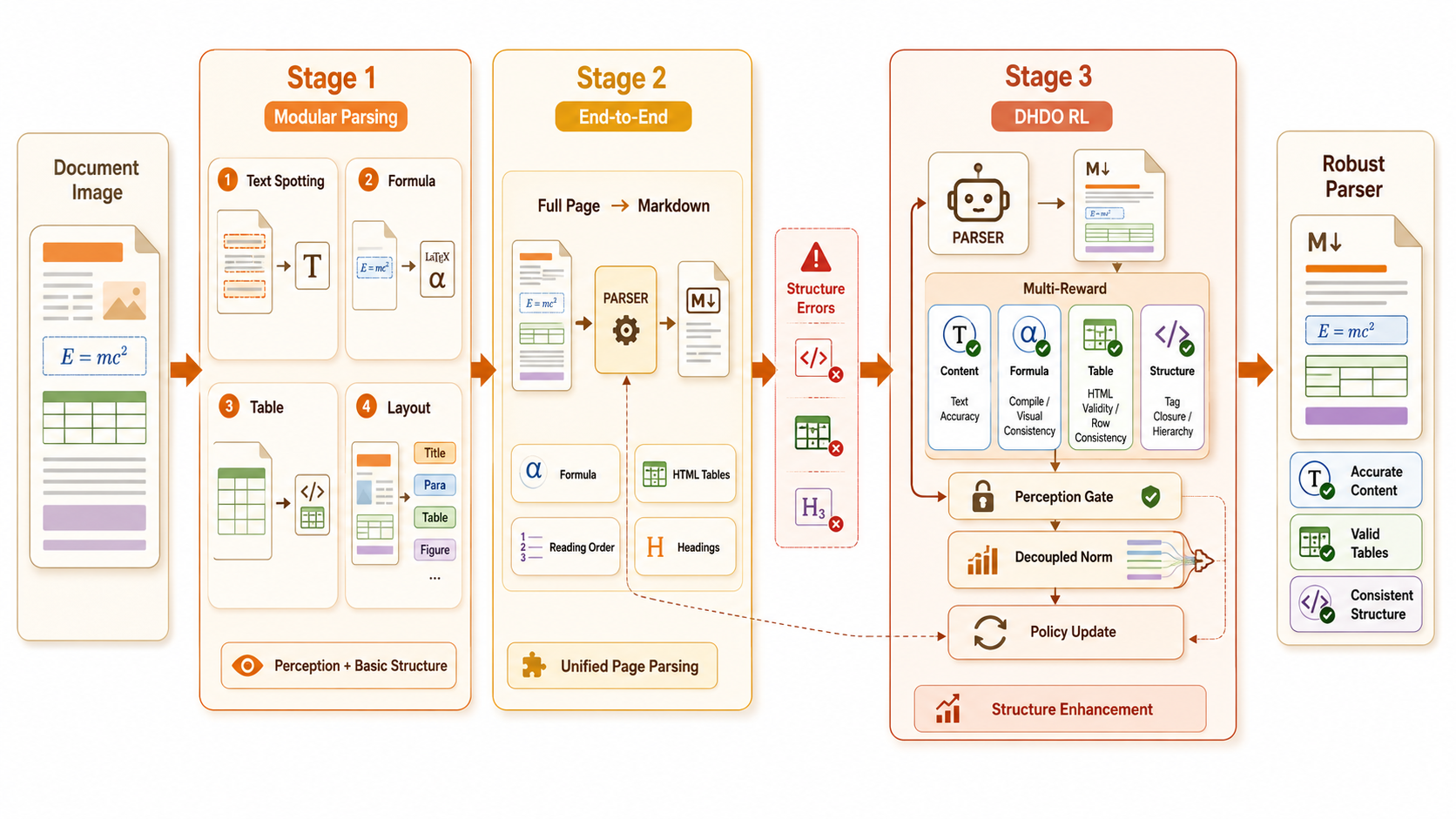}
    \caption{\textbf{The three-stage training pipeline.} The framework advances document parsing through: (1) \textbf{Modular foundation}: building capabilities in text, formula, table, and layout; (2) \textbf{End-to-end unification}: direct page-to-Markdown generation; (3) \textbf{Structural enhancement}: refining output quality via structure-constrained reinforcement learning, incorporating perception-aware and structure-aware rewards with decoupled normalization.}
    \label{fig:three_stage_pipeline}
\end{figure}

We adopt Qwen3-VL-2B-Instruct \cite{qwen3vl2025} as our backbone model. As illustrated in \Cref{fig:three_stage_pipeline}, Stage 1 and Stage 2 enhance the perception and structural capabilities via cross-entropy supervision. Specifically, we decompose document parsing into distinct sub-tasks: Stage 1 targets four specific sub-capabilities, while Stage 2 focuses on end-to-end document parsing. As cross-entropy offers weak supervision for document structure due to its limitations with long-range dependencies, Stage 3 utilizes structure-constrained reinforcement learning to further enhance structural capabilities.

\subsection{Stage 1: Modular Document Parsing}
\label{sec:stage1}

In Stage 1, we decompose the complex end-to-end document parsing process into four sub-tasks. Proficiency in these foundational capabilities is pivotal to the success of the final end-to-end document parsing.

\paragraph{Text spotting}
Given a full-page document image, the model is supposed to precisely locate all content blocks while accurately recognizing their content, adhering to the correct reading order.
\[
  \texttt{<|box\_start|>(x1,y1),(x2,y2)<|box\_end|>text\_content},
\]

\paragraph{Formula recognition} Given an image containing a single formula, or a full-page document image with a region indicator (e.g., a bounding box), the model should to parse the formula into a normalized \LaTeX{} format, correctly handling inline, display, and multi-line expressions.

\paragraph{Table recognition}
Given an image containing an isolated table, the model is expected to correctly parse the table in the form of normalized HTML.
\[
    \verb|<table><tr><td>...</td></tr></table>|.
\]

\paragraph{Layout Analysis}
Given a full-page document image, the model is supposed to generate a sequence of region boxes, each paired with a semantic category from the set
\{\textit{title, text, table, figure, formula, footer, $\dots$}\}.
\[
  \texttt{<|box\_start|>(x1,y1),(x2,y2)<|box\_end|><type>title</type>}.
\]

\medskip

The training dataset in Stage~1 is the union of four distinct training sets of each sub-task: 
$\mathcal{D}_{\text{S1}} = \mathcal{D}_{\text{spot}} \cup \mathcal{D}_{\text{formula}} \cup \mathcal{D}_{\text{table}} \cup \mathcal{D}_{\text{layout}}$.
The model is optimized through minimizing cross-entropy loss.
\begin{equation}
  \mathcal{L}_{\text{S1}}(\theta)
  = -\mathbb{E}_{(I, c, y) \sim \mathcal{D}_{\text{S1}}}
    \left[
      \sum_{t=1}^{|y|} \log \pi_{\theta}(y_t \mid y_{<t}, I, c)
    \right],
  \label{eq:l-stage1}
\end{equation}
where $c$ denotes the task-specific prompt.

These four sub-tasks synergistically enhance the model's perception and structural capabilities from diverse perspectives. For instance, text spotting simultaneously boosts the model's OCR performance (perception capability) and its ability to segment blocks according to the layout (structural capability). Following the first stage of training, the model establishes a robust foundation in terms of both perception and structural capabilities.

\subsection{Stage 2: End-to-End Document Parsing}
\label{sec:stage2}

Stage 2 is specifically designed to boost the structural and perception capabilities through end-to-end training on page-level document parsing data. To maintain structural consistency, we introduce specific data processing steps and experimental setups. Let $\mathcal{D}_{\text{S2}} = \{(I^{\text{page}}_i, M^{\star}_i)\}_{i=1}^{N_2}$.
Ground-truth Markdown $M^{\star}$ is obtained through a strict normalization pipeline with the following invariants:
\begin{itemize}[leftmargin=1.4em, itemsep=0.15em, topsep=0.3em]
  \item Display mathematics uses \verb|$$ ... $$|; inline mathematics uses \verb|$ ... $|.
  \item All tables are represented in HTML, including those with merged cells.
  \item Reading order follows visual layout: on multi-column pages, columns are emitted left to right, with each column completed before advancing.
  \item Heading depth is encoded by \verb|#|, \verb|##|, and \verb|###| without skipped levels (e.g., no transition from \verb|#| directly to \verb|###|).
\end{itemize}

The training objective of stage 2 is: 
\begin{equation}
  \mathcal{L}_{\text{S2}}(\theta)
  = -\mathbb{E}_{(I^{\text{page}}, M^{\star}) \sim \mathcal{D}_{\text{S2}}}
    \left[
      \sum_{t=1}^{|M^{\star}|}
        \log \pi_{\theta}\!\left(
          M^{\star}_t \mid M^{\star}_{<t}, I^{\text{page}}
        \right)
    \right].
  \label{eq:l-stage2}
\end{equation}

The perception capability is supposed to be maximized after Stage 2, allowing for the complete identification of content in page-level document images. Nevertheless, cross-entropy offers weak supervision for document structure, largely due to its limitations with long-range dependencies. As a result, structural inaccuracies persist, including issues like unclosed tags, inconsistent table row widths, and occasional heading-level skips.

\subsection{Stage 3: Structure-Constrained Reinforcement Learning}
\label{sec:stage3}

While the cross-entropy supervision in both Stage 1 and Stage 2 effectively enhances the perception capabilities, it does not impose strong constraints on heterogeneous document structuring, such as Markdown-compliant tables and LaTeX-formatted formulas. To ensure robust document structuring, prior approaches \cite{hunyuanocr, fireredocr2025} adopt GRPO \citep{shao2024deepseekmath}, incorporating various rewards corresponding to different document components. 

However, prior studies \cite{liu2026gdpo} have revealed that naively applying GRPO to normalize diverse reward combinations across a wide range of rewards (including accuracy, format correctness, length constraints, and code quality) leads to an advantage collapse, where distinct signals are mapped to identical values, thereby degrading the learning signal and hindering optimal convergence. We analyze that this collapse also stems from the fact that the causal dependencies among rewards are neglected. Similarly, utilizing GRPO in document parsing would inevitably neglect the causality between heterogeneous rewards.

To address this challenge, we draw upon the insights from Group reward-Decoupled Normalization
Policy Optimization (GDPO) \cite{liu2026gdpo}, which mitigates advantage collapse by decoupling the normalization process. Specifically, this approach first performs group-wise normalization on each reward component independently to preserve their relative distinctions. These normalized advantages are then summed to compute the overall advantage, followed by a final batch-wise normalization to ensure numerical stability. By maintaining the granularity of diverse reward signals, this approach effectively prevents signal degradation. Building upon this foundation, we propose a novel reinforcement learning framework for document parsing: \textbf{Decoupled Heterogeneous Document Optimization (DHDO)}.

\subsubsection{Reward Designs}
\label{sec:rewards}

The goal of DHDO is to strengthen structuring constraints without compromising perception capabilities. We introduce separate perception and structure rewards, recognizing the inherent causality that structuring correctness should depend on accurate perception.

\paragraph{(1) Perception Reward - Content Accuracy} Levenshtein similarity is employed to measure content accuracy. Elements in Markdown and HTML that do not affect visual consistency (e.g., whitespace) are stripped prior to evaluation. :
\begin{equation}
  \Rtext(\hat{y}, y^{\star})
  = 1 - \frac{
      \Lev\!\left(\Pi(\hat{y}),\, \Pi(y^{\star})\right)
    }{
      \max\!\left(|\Pi(\hat{y})|,\, |\Pi(y^{\star})|\right)
    },
  \label{eq:r-text}
\end{equation}
where $\Pi(\cdot)$ denotes the strip-markup projection.

\paragraph{(2) Structure Reward - Formula Syntactic Soundness}
For each predicted formula $\hat{f}_k$,  we verify the soundness under KaTeX and evaluate its visual equivalence to the corresponding reference using Character Detection Matching (CDM) \citep{unimernet2024}:
\begin{equation}
  \Rformula(\hat{y}, y^{\star})
  = \frac{1}{K} \sum_{k=1}^{K}
    \Big[
      \alpha\, \mathbb{1}\!\left[\texttt{compile}(\hat{f}_k)\right]
      + (1 - \alpha)\, \mathrm{CDM}(\hat{f}_k, f^{\star}_k)
    \Big],
  \label{eq:r-formula}
\end{equation}
where we set $\alpha = 0.3$. A success compilation is a necessary but insufficient condition for semantic correctness. Therefore, to account In contrast, CDM aligns more closely with human-perceived visual quality. By convention, we define $\Rformula = 1$ when neither $\hat{y}$ nor $y^{\star}$ contains a formula. 

\paragraph{(3) Structure Reward - Table Structural Validity}
Given that the training data is represented in HTML, we define the reward, $\mathcal{R}_{\text{table}}$, directly on the HTML structure. This reward function is formulated as a weighted combination of structural shape constraints and tree-edit similarity:
\begin{equation}
  \mathcal{R}_{\text{table}}(\hat{y}, y^{\star})
  = \beta\, S_{\text{shape}}^{\text{HTML}}(\hat{y}, y^{\star})
    + (1 - \beta)\, \mathrm{TEDS}(\hat{y}, y^{\star}),
  \label{eq:r-table}
\end{equation}
where $\mathrm{TEDS}$ denotes the Tree-Edit-Distance-based Similarity \citep{zhong2020image} computed on the parsed table trees. The shape consistency term, $S_{\text{shape}}^{\text{HTML}}$, factorizes into three distinct validation checks:
\begin{equation}
  S_{\text{shape}}^{\text{HTML}}(\hat{y}, y^{\star})
  = \underbrace{\mathbb{1}\!\left[\text{well-formness}(\hat{y})\right]}_{\text{valid tag nesting}}
    \cdot
    \underbrace{\mathbb{1}\!\left[\text{cells-consistency}(\hat{y})\right]}_{\text{uniform row widths}}
    \cdot
    \underbrace{\exp\!\left(-\gamma\, \left|N_{\text{row}}(\hat{y}) - N_{\text{row}}(y^{\star})\right|\right)}_{\text{row-count alignment}}.
  \label{eq:r-table-shape}
\end{equation}

The \textit{well-formedness} check is enforced by parsing $\hat{y}$ using \texttt{lxml.html} in strict mode; any parsing failure results in a zero reward contribution. 
The \textit{cells-consistency} check ensures structural uniformity by computing the effective row width $W(r) = \sum_{c \in \text{cells}(r)} \texttt{colspan}(c)$, accounting for active \texttt{rowspan} spans, and requiring $W(r) = W(r')$ for all row pairs $(r,r')$. 
This criterion offers a balanced constraint: it is more robust than naive column counting when handling merged cells, yet stricter than simple row-count checks as it implicitly rejects truncated rows. Throughout our experiments, we set the hyperparameters to $\beta = 0.4$ and $\gamma = 0.1$. Conventionally, if $\hat{y}$ lacks a \texttt{<table>} tag while $y^{\star}$ contains one, $\mathcal{R}_{\text{table}}$ is set to 0; conversely, if neither contains a table, $\mathcal{R}_{\text{table}}$ defaults to 1.

\paragraph{(4) Structure Reward - General Structural Closure}
Let $\mathcal{T}$ denote the set of paired Markdown delimiters and HTML table tags, formally defined as:
\(
\mathcal{T} = \{\texttt{**}, \texttt{\_\_}, \texttt{\$\$}, \texttt{`}\}
\cup \{\texttt{<table>}, \texttt{<thead>}, \texttt{<tbody>}, \texttt{<tr>}, \texttt{<td>}, \texttt{<th>}\}.
\)
The structural closure reward is then formulated as:
\begin{equation}
  \mathcal{R}_{\text{struct}}(\hat{y})
  = \left[
      1 - \frac{1}{|\mathcal{T}|}
      \sum_{\tau \in \mathcal{T}}
        \frac{
          \left|N_{\text{open}}(\tau; \hat{y}) - N_{\text{close}}(\tau; \hat{y})\right|
        }{
          N_{\text{open}}(\tau; \hat{y}) + N_{\text{close}}(\tau; \hat{y}) + \epsilon
        }
    \right]
    \cdot \mathbb{1}\!\left[\text{hierarchy-valid}(\hat{y})\right],
  \label{eq:r-struct}
\end{equation}
where the multiplicative indicator function enforces strict heading hierarchy constraints (e.g., prohibiting level skips such as \verb|#| to \verb|###|) and valid list indentation.  Notably, $\mathcal{R}_{\text{struct}}$ and $\mathcal{R}_{\text{table}}$ are designed to be complementary. While $\mathcal{R}_{\text{struct}}$ is a computationally inexpensive and coarse-grained metric, it provides crucial early reinforcement learning signals during the initial training stages, particularly when the more complex $\mathcal{R}_{\text{table}}$ reward remains near zero across most rollouts.

We introduce a perception reward and three structure rewards designed for heterogeneous document structures. However, these objectives exhibit significant disparities in difficulty. To prevent reward hacking, we condition the structure rewards on the perception reward, making accurate perception a strict prerequisite for structural optimization.

\begin{equation}
  \tilde{R}_{\bullet}
  = R_{\bullet}\, \mathbb{1}\!\left[\Rtext \geq \tau_{\text{text}}\right],
  \qquad \tau_{\text{text}} = 0.7.
  \label{eq:cond-reward}
\end{equation}

\subsubsection{Reward Aggregation and Policy Optimization}

For each prompt $I^{(i)}$, $G$ rollouts $\{o^{(i,j)}\}_{j=1}^{G}$ are sampled from $\pi_{\theta_{\text{old}}}$, and the four (possibly conditioned) rewards $r_k^{(i,j)}$ are evaluated for
$k \in \{\text{text, formula, table, struct}\}$.
Each reward dimension is first standardized within the group, then aggregated across dimensions, and finally rescaled over the minibatch $\mathcal{B}$:
\begin{equation}
  A_k^{(i,j)} = \frac{r_k^{(i,j)} - \mu_k^{(i)}}{\sigma_k^{(i)} + \epsilon},
  \qquad
  A_{\text{sum}}^{(i,j)} = \sum_k w_k\, A_k^{(i,j)},
  \qquad
  \hat{A}^{(i,j)} = \frac{A_{\text{sum}}^{(i,j)} - \mu_{\mathcal{B}}}{\sigma_{\mathcal{B}} + \epsilon},
  \label{eq:gdpo-adv}
\end{equation}
where $\mu_k^{(i)}$ and $\sigma_k^{(i)}$ are the within-group mean and standard deviation of reward $k$.
Per-reward normalization retains gradient signal that would be lost if rewards were summed before normalization; batch-level rescaling stabilizes the effective learning rate as the reward dimensionality changes.
Aggregation weights are set to
$w_{\text{text}} : w_{\text{formula}} : w_{\text{table}} : w_{\text{struct}} = 1.0 : 0.8 : 0.8 : 0.5$,
prioritizing perception accuracy over structure constraints, and aligning with the conditioning gate in \eqref{eq:cond-reward}.

The policy is updated by maximizing the clipped surrogate \citep{schulman2017proximal} with $\hat{A}^{(i,j)}$ substituted for the usual advantage, regularized by
$\beta_{\KL}\, \KL(\pi_{\theta} \,\|\, \pi_{\text{ref}})$,
which anchors $\pi_{\theta}$ to the frozen Stage~2 checkpoint $\pi_{\text{ref}}$.

\section{Experiments}

We present a systematic evaluation of ABot-OCR. To establish a comprehensive benchmark, we compare our model against three representative categories of methods:
\begin{itemize}[leftmargin=1.4em, itemsep=0.2em, topsep=0.3em]
\item \textbf{General Vision-Language Models}: This group includes massive general VLMs such as Qwen3-VL-235B~\cite{qwen3vl235b} and Gemini-3.0 Pro~\cite{gemini3pro}.
\item \textbf{Multi-Stage Pipeline Systems}: This category comprises traditional modular pipelines, represented by systems like PaddleOCR-VL-1.5~\cite{paddleocr_vl15}.
\item \textbf{Specialized End-to-End OCR Models}: This benchmark includes dedicated document-parsing architectures such as DeepSeek-OCR 2~\cite{deepseek_ocr2} and dots.ocr~\cite{dots_ocr}.
\end{itemize}

\subsection{Dataset}

For real-world document OCR and layout parsing, our data construction strategy focuses on two core goals: wide coverage and aligned supervision. In addition to high-quality open-source corpora, we re-annotate selected public datasets. This re-annotation ensures that all supervision aligns with a single, unified objective: mapping images directly to structured text, including complex formulas and tables. This process introduces massive variety in layouts, typefaces, and domains, which heavily boosts our model's generalizability.

To build specialized capabilities, we organize our training data into five task-specific categories:
\begin{itemize}[leftmargin=1.4em, itemsep=0.2em, topsep=0.3em]
\item \textbf{Handwriting Recognition}: We use the standard CASIA-HWDB~\cite{casia_hwdb} dataset. We convert all annotations into Markdown format to perfectly match our full-document parsing pipeline.
\item \textbf{Table Understanding}: We adopt the widely used PubTabNet~\cite{pubtabnet} dataset to train the model on tabular structures.
\item \textbf{Mathematical Expressions}: We combine multiple rich resources, including UniMER-1M~\cite{unimernet2024}, MathWriting~\cite{mathwriting2024}, LaTeX OCR~\cite{blecher2022latexocr}, and latex-formulas-80M~\cite{latex_formulas_80m}. Because these open-source datasets use inconsistent \LaTeX{} styles (such as variations in spacing and symbols), we develop an expression normalization pipeline during preprocessing. This step standardizes all \LaTeX{} strings and eliminates the syntactic gap between training and evaluation.
\item \textbf{Chart Understanding}: To boost the model's performance on charts, we integrate a diverse set of chart resources. These include ChartQA~\cite{chartqa2022}, PlotQA~\cite{plotqa2020}, Chart2Text~\cite{chart2text2022}, DVQA~\cite{dvqa2018}, Unichart~\cite{unichart2023}, Beagle~\cite{battle2018beagle}, Chart-INFO~\cite{chartinfo2024}, visText~\cite{vistext2023}, and ExcelChart~\cite{excelchart2021}. During integration, we apply strict data cleaning to filter out low-quality samples and remove duplicate entries.
\item \textbf{General Document Parsing}: We construct a large-scale web document corpus. This dataset covers a wide array of document types and visual styles, including academic papers, newspapers, journal articles, scanned forms, e-books, exam papers, and presentation slides. Mixing these heterogeneous sources prevents the model from overfitting to clean, standard layouts.
\end{itemize}

To further patch the model's weaknesses in extreme or rare scenarios, we also use a targeted data synthesis pipeline. Specifically, we combine extensive font libraries, diverse CSS styles, and multilingual corpora. We then render these components into hard but high-quality training samples using XeLaTeX and modern web browsers.

Ultimately, this carefully curated data mixture provides layered support for both standard text reading and specialized structural parsing, establishing a rock-solid data foundation for ABot-OCR.

\subsection{Training Recipe}

We train our model using a structured, three-stage schedule. The key hyperparameters for each stage are summarized in Table~\ref{tab:train_three_stage}.

During the first two stages, we maintain consistency by sharing the same global batch size and base learning rate settings. In the third stage, we switch to a structure-constrained DHDO training strategy, where we reduce the learning rate to fine-tune the model. For the decoding process in this final stage, we implement nucleus sampling. Additionally, we set predefined budgets for both the context length and the generation length to ensure efficient inference.

\begin{table}[htbp]
  \centering
  \caption{Three-stage training configuration.}
  \label{tab:train_three_stage}
  \small
  \begin{tabularx}{\textwidth}{@{}l Y r c c Y@{}}
    \toprule
    Stage & Objective & Data scale & LR & GBS & Others \\
    \midrule
    Modular Parsing &
    Strengthen visual representations and multitask foundations &
    $\sim$10\,M &
    $5\times10^{-5}$ &
    128 &
    Linear warmup ratio $0.05$ \\
    
    End-to-End &
    Image$\rightarrow$Markdown; enforce structured outputs &
    $\sim$1.4\,M &
    $5\times10^{-5}$ &
    128 &
    Same as above \\
    
    DHDO &
    Policy refinement; format constraints and stability &
    $\sim$200\,k &
    $5\times10^{-7}$ &
    128 &
    Nucleus sampling: $p=0.99$, $k=50$ \\
    \bottomrule
  \end{tabularx}
\end{table}

\subsection{Evaluation Results}

We evaluate our model on two standard benchmarks: OmniDocBench v1.5~\cite{omnidocbench2024} and OmniDocBench v1.6~\cite{mineru25pro}. To provide a thorough analysis, we report both the Overall Score and several fine-grained evaluation metrics. First, we use the standard Edit Distance to measure text transcription accuracy and reading-order prediction, where a lower score indicates better performance. Second, we employ the CDM metric to evaluate the structural and semantic consistency of mathematical expressions. Finally, we use the TEDS score to evaluate the quality of table structure reconstruction.

\subsubsection{Overall Analysis on OmniDocBench v1.5}

As summarized in Table~\ref{tab:omnires1.5}, we benchmark ABot-OCR against three main types of document parsers on the OmniDocBench v1.5 dataset. These include multi-stage pipeline systems, large general Vision-Language Models (VLMs), and specialized end-to-end (E2E) OCR models. Despite its compact 2B parameter scale, ABot-OCR remains highly competitive across all major evaluation tasks, including text transcription, formula parsing, table structure reconstruction, and reading-order recovery. These results prove that ABot-OCR achieves an exceptional balance between parsing accuracy, structural fidelity, and low deployment complexity. Therefore, it is perfectly suited for real-world document understanding scenarios.

\textit{Comparison with specialized end-to-end OCR models.}
Within the specialized E2E OCR category, ABot-OCR achieves the top performance. It delivers the best Overall score in this group with a 92.81, successfully surpassing the strong same-scale baseline FireRed-OCR, which scores 92.07. Furthermore, ABot-OCR demonstrates clear advantages over larger end-to-end models like DeepSeek-OCR 2 (3B) and dots.ocr (3B). Our model wins not only in the Overall score but also in the fine-grained character recognition metric. Specifically, ABot-OCR lowers the TextEdit error to a mere 0.034, which is significantly better than DeepSeek-OCR's 0.049 and dots.ocr's 0.048.

\textit{Performance relative to large general VLMs.}
A key takeaway from our evaluation is that document parsing performance does not automatically grow just by scaling up a generic model. For example, the massive Qwen3-VL-235B model only achieves an Overall score of 89.15 and a TextEdit score of 0.069. Both of these scores are substantially worse than ABot-OCR's performance of 92.81 and 0.034. This comparison proves that domain-specific optimization and task alignment are far more critical for structure-intensive document parsing than pure parameter size. As a result, a compact, OCR-specialized model can deliver much stronger parsing accuracy while requiring only a fraction of the compute and maintenance costs.

\textit{Robustness relative to pipeline systems.}
As Table~\ref{tab:omnires1.6} shows, traditional multi-stage pipeline systems can still reach a slightly higher performance ceiling by stacking independent sub-modules. For instance, PaddleOCR-VL-1.5 (0.9B) achieves an Overall score of 94.50, outperforming our model. However, these small performance gains come with a massive cost. Pipeline systems suffer from heavy engineering overhead, including multi-stage execution, complex module coordination, and fragile software version dependencies. In contrast, our end-to-end architecture enables unified deployment and a vastly simplified serving pipeline. At the same time, ABot-OCR achieves a TextEdit score of 0.034, which is slightly better than PaddleOCR-VL-1.5's 0.035. This result confirms that our end-to-end model does not sacrifice character-level accuracy. The remaining gap in the Overall score comes mostly from specialized structural sub-tasks, like formulas and tables, where pipeline systems still hold a narrow advantage.

\subsubsection{Fine-grained Analysis on OmniDocBench v1.5}

\textit{Intra-family structural improvements.}
Compared to FireRed-OCR, ABot-OCR improves both table-related metrics at the same time. Specifically, the TEDS score increases from 88.72 to 90.45, while the TEDS\textsubscript{s} score rises from 92.38 to 93.96. These gains align perfectly with the improvement in our Overall score. They prove that ABot-OCR learns much better internal representations for table topologies and cell alignments. Additionally, ABot-OCR maintains a clear advantage on the formula metric, achieving a higher CDM score of 91.38 compared to 90.98 for FireRed-OCR.

\textit{Advantages over larger end-to-end models.}
When compared to DeepSeek-OCR 2 (3B), our smaller ABot-OCR delivers substantially better performance on both the Overall and TextEdit metrics. Furthermore, it widens the gap even more in table reconstruction tasks. For instance, ABot-OCR achieves a TEDS score of 90.45 and a TEDS\textsubscript{s} score of 93.96, while DeepSeek-OCR 2 only scores 85.60 and 90.06 respectively. This clear trend shows that table structure reconstruction does not automatically improve just by making an end-to-end model larger. Instead, it relies heavily on task-aligned data composition, structural supervision, and carefully designed training recipes.

\textit{Text fidelity and reading-order consistency.}
ABot-OCR matches FireRed-OCR on the reading-order metric with a score of 0.041. This ties both models for the absolute best performance among all current end-to-end systems. In contrast, Qwen3-VL-235B scores a much higher 0.068 on the same metric. This comparison highlights that ABot-OCR dramatically lowers reading-order error rates, proving that ordering mistakes are no longer a major failure mode in our system. Meanwhile, ABot-OCR achieves a TextEdit score of 0.034, which is slightly better than FireRed-OCR's 0.035. This result confirms our model's superior character recognition capability at the 2B parameter scale.

\begin{table}[htbp]
\centering
\caption{Performance comparison of document parsing methods on OmniDocBench v1.5.}
\label{tab:omnires1.5}
\small
\setlength{\tabcolsep}{4pt}
\renewcommand{\arraystretch}{1.1}

\begin{tabular}{@{}l l S[table-format=2.2] S[table-format=1.3] S[table-format=2.2] S[table-format=2.2] S[table-format=2.2] S[table-format=1.3]@{}}
\toprule
\textbf{Methods} & \textbf{Param} & 
\multicolumn{1}{c}{\textbf{Overall↑}} & 
\multicolumn{1}{c}{\textbf{TextEdit↓}} & 
\multicolumn{1}{c}{\textbf{Formula\textsuperscript{CDM}↑}} & 
\multicolumn{1}{c}{\textbf{Table\textsuperscript{TEDS}↑}} & 
\multicolumn{1}{c}{\textbf{Table\textsuperscript{TEDS\(_s\)}↑}} & 
\multicolumn{1}{c}{\textbf{R-order\textsuperscript{Edit}↓}} \\
\midrule

\multicolumn{8}{@{}l}{Pipeline OCR Systems} \\
PaddleOCR-VL-1.5 ~\cite{paddleocr_vl15} & 0.9B & \textbf{94.50} & \textbf{0.035} & \textbf{94.21} & 92.76 & 95.79 & \underline{0.042} \\
GLM-OCR ~\cite{glmocr}  & 0.9B & \underline{94.35} & 0.045 & \underline{93.65} & \textbf{93.89} & \textbf{96.50} & 0.047 \\
Youtu-Parsing ~\cite{youtu_parsing} & 2.5B & 93.37 & 0.042 & 91.22 & \underline{93.10} & \underline{96.47} & \textbf{0.026} \\
PaddleOCR-VL ~\cite{paddleocr_vl} & 0.9B & 92.86 & \textbf{0.035} & 91.22 & 90.89 & 94.76 & 0.043 \\
Logics-Parsing-v2 ~\cite{logics_parsing_v2} & 4B & 92.56 & 0.043 & 91.41 & 90.54 & 93.85 & 0.044 \\
MinerU2.5 ~\cite{mineru25} & 1.2B & 90.93 & 0.045 & 88.86 & 88.44 & 92.42 & 0.044 \\
MonkeyOCR-pro-3B ~\cite{monkeyocr_pro_3b} & 3B & 88.85 & 0.075 & 87.25 & 86.78 & 90.63 & 0.128 \\
Dolphin-v2 ~\cite{dolphin_v2} & 3B & 88.71 & 0.073 & 87.26 & 86.20 & 89.77 & 0.064 \\

\midrule
\addlinespace
\multicolumn{8}{@{}l}{End-to-End OCR Models} \\
\textbf{ABot-OCR(Ours)} & 2B & \textbf{92.81} & \textbf{0.034} &  \textbf{91.38} & \underline{90.45} & \underline{93.96} & \textbf{0.041} \\
FireRed-OCR ~\cite{fireredocr2025} & 2B & \underline{92.07} & \underline{0.035} & \underline{90.98} & 88.72 & 92.38 & \underline{0.041} \\
HunyuanOCR ~\cite{hunyuanocr}  & 1B & 90.57 & 0.085 & 86.01 & \textbf{94.19} & \textbf{95.96} & 0.082 \\
OpenDoc-0.1B ~\cite{unirec01b} & 0.1B & 90.57 & 0.043 & 87.70 & 88.30 & 92.24 & 0.050 \\
DeepSeek-OCR 2 ~\cite{deepseek_ocr2} & 3B & 89.17 & 0.049 & 86.85 & 85.60 & 90.06 & 0.060 \\
OCRVerse  ~\cite{ocrverse} & 4B & 88.55 & 0.058 & 86.91 & 84.55 & 88.45 & 0.071 \\
dots.ocr ~\cite{dots_ocr} & 3B & 88.41 & 0.048 & 83.22 & 86.78 & 90.62 & 0.053 \\

\midrule
\addlinespace
\multicolumn{8}{@{}l}{General VLMs} \\
Ovis2.6-30B-A3B ~\cite{ovis25report} & 30B & 92.36 & 0.037 & 90.32 & 90.46 & 94.00 & 0.046 \\
Gemini 3 Flash \faLock & – & 90.37 & 0.065 & 89.56 & 88.01 & 93.79 & 0.071 \\
Gemini 3 Pro \faLock & – & 90.17 & 0.062 & 88.79 & 87.83 & 93.32 & 0.074 \\
Qwen3-VL-235B ~\cite{qwen3vl235b}  & 235B & 89.15 & 0.069 & 88.14 & 86.21 & 90.55 & 0.068 \\
GPT-5.2 \faLock & –  & 85.75 & 0.124 & 86.93 & 82.76 & 88.25 & 0.106 \\
InternVL3.5-241B ~\cite{internvl35}  & 241B & 82.67 & 0.142 & 87.23 & 75.00 & 81.28 & 0.125 \\

\bottomrule
\end{tabular}
\end{table}

\subsubsection{Analysis on OmniDocBench v1.6}
To further verify our model on the premium standard of document parsing, we evaluate its performance on the latest OmniDocBench v1.6 benchmark~\cite{mineru25pro}. Table~\ref{tab:omnires1.6} presents the detailed comparative results.

Currently, modular pipeline parsers still maintain the highest Overall scores. This advantage is expected because they combine multiple independent, highly specialized modules for layout detection and text recognition. However, these traditional pipelines require complex, multi-stage coordination. In contrast, our end-to-end results prove that a single 2B model can achieve a nearly identical level of text fidelity, completely avoiding the engineering overhead of multi-stage orchestration.

When looking specifically at the end-to-end OCR models, ABot-OCR achieves the top performance. It delivers the best Overall score (93.30) and the lowest TextEdit distance (0.037), tying for first place with FireRed-OCR. At the same time, it secures the second-highest table metrics with a TEDS score of 88.83 and a TEDS\textsubscript{s} score of 91.94. This outstanding performance proves that ABot-OCR maintains an excellent balance between fine-grained text transcription and complex tabular structure reconstruction, all within a single forward pass.

Furthermore, we compare our model against general-purpose Vision-Language Models (VLMs). Compared to the same-scale Qwen3-VL-2B, ABot-OCR shows a massive performance lead across all reported metrics. More importantly, ABot-OCR clearly outperforms substantially larger models, such as the Qwen3-VL-235B, in both Overall scores and TextEdit accuracy. This crucial comparison strongly supports our core argument: parameter-efficient, parsing-oriented training is far more effective for document understanding than simply relying on brute-force parameter scaling.

\begin{table}[htbp]
\centering
\caption{Performance comparison of document parsing methods on OmniDocBench v1.6 Full across text, formula, table, and reading order extraction tasks.}
\label{tab:omnires1.6}
\small
\setlength{\tabcolsep}{4pt}
\renewcommand{\arraystretch}{1.1}

\begin{tabular}{@{}l l S[table-format=2.2] S[table-format=1.3] S[table-format=2.2] S[table-format=2.2] S[table-format=2.2] S[table-format=1.3]@{}}
\toprule
\textbf{Methods} & \textbf{Param} & 
\multicolumn{1}{c}{\textbf{Overall↑}} & 
\multicolumn{1}{c}{\textbf{TextEdit↓}} & 
\multicolumn{1}{c}{\textbf{Formula\textsuperscript{CDM}↑}} & 
\multicolumn{1}{c}{\textbf{Table\textsuperscript{TEDS}↑}} & 
\multicolumn{1}{c}{\textbf{Table\textsuperscript{TEDS\(_s\)}↑}} & 
\multicolumn{1}{c}{\textbf{R-order\textsuperscript{Edit}↓}} \\
\midrule

\multicolumn{8}{@{}l}{Pipeline OCR Systems} \\
\textbf{MinerU2.5-Pro} ~\cite{mineru25pro} & 1.2B & \textbf{95.69} & \textbf{0.036} & \textbf{97.29} & \textbf{93.42} & \textbf{95.92} & \underline{0.120} \\
GLM-OCR ~\cite{glmocr}  & 0.9B & \underline{95.15} & 0.044 & \underline{96.99} & \underline{92.83} & \underline{95.39} & 0.133 \\
PaddleOCR-VL-1.5 ~\cite{paddleocr_vl15} & 0.9B & 94.87 & \underline{0.038} & 96.69 & 91.67 & 94.37 & 0.130 \\
PaddleOCR-VL ~\cite{paddleocr_vl} & 0.9B & 94.11 & 0.040 & 95.70 & 90.65 & 93.74 & 0.135 \\
Youtu-Parsing ~\cite{youtu_parsing} & 2.5B & 93.68 & 0.044 & 93.45 & 92.02 & 95.00 & \textbf{0.116} \\
Logics-Parsing-v2 ~\cite{logics_parsing_v2} & 4B & 93.27 & 0.041 & 95.47 & 88.42 & 91.98 & 0.137 \\
MinerU2.5 ~\cite{mineru25} & 1.2B & 92.98 & 0.045 & 95.59 & 87.88 & 91.47 & 0.130 \\
Dolphin-v2 ~\cite{dolphin_v2} & 3B & 89.34 & 0.069 & 90.53 & 84.40 & 87.44 & 0.150 \\
MonkeyOCR-pro-3B ~\cite{monkeyocr_pro_3b} & 3B & 88.43 & 0.074 & 88.33 & 84.35 & 88.62 & 0.189 \\

\midrule
\addlinespace
\multicolumn{8}{@{}l}{End-to-End OCR Models} \\
\textbf{ABot-OCR(Ours)} & 2B & \textbf{93.30} & \textbf{0.037} &  \underline{94.86} & \underline{88.83} & \underline{91.94} & \underline{0.133} \\
FireRed-OCR ~\cite{fireredocr2025} & 2B & \underline{93.20} & \textbf{0.037} & \textbf{95.27} & 88.04 & 91.06 & \textbf{0.131} \\
OpenDoc-0.1B ~\cite{unirec01b} & 0.1B & 90.64 & 0.049 & 92.93 & 83.88 & 87.45 & 0.140 \\
dots.ocr ~\cite{dots_ocr} & 3B & 90.50 & 0.048 & 89.12 & 87.18 & 90.58 & 0.138 \\
DeepSeek-OCR 2 ~\cite{deepseek_ocr2} & 3B & 90.17 & 0.050 & 91.59 & 83.89 & 87.75 & 0.144 \\
HunyuanOCR ~\cite{hunyuanocr}  & 1B & 89.87 & 0.089 & 87.44 & \textbf{91.01} & \textbf{93.23} & 0.171 \\
OCRVerse  ~\cite{ocrverse} & 4B & 88.44 & 0.063 & 89.14 & 82.44 & 86.27 & 0.163 \\

\midrule
\addlinespace
\multicolumn{8}{@{}l}{General VLMs} \\
Ovis2.6-30B-A3B ~\cite{ovis25report} & 30B & 93.62 & 0.035 & 94.93 & 89.44 & 92.40 & 0.135 \\
Gemini 3 Pro \faLock & – & 92.85 & 0.064 & 95.83 & 89.15 & 92.96 & 0.165 \\
Gemini 3 Flash \faLock & – & 92.58 & 0.066 & 95.03 & 89.29 & 93.51 & 0.173 \\
Qwen3-VL-2B ~\cite{qwen3vl235b} & 2B & 80.05 & 0.091 & 80.36 & 68.94 & 73.82 & 0.213 \\
Qwen3-VL-235B ~\cite{qwen3vl235b}  & 235B & 89.78 & 0.063 & 92.53 & 83.07 & 86.75 & 0.166 \\
GPT-5.2 \faLock & –  & 86.52 & 0.114 & 88.00 & 82.95 & 87.93 & 0.193 \\
InternVL3.5-241B ~\cite{internvl35}  & 241B & 83.61 & 0.130 & 89.52 & 74.35 & 79.78 & 0.215 \\

\bottomrule
\end{tabular}
\end{table}

\subsubsection{Performance on Multilingual OCR}

To evaluate cross-lingual generalizability, we train and evaluate our model on an in-house multilingual OCR dataset covering 10 diverse languages. Table~\ref{tab:multilingual_transposed} reports the text edit distances for all evaluated systems.

The empirical results show that ABot-OCR achieves the best overall performance, yielding a remarkably low average edit distance of \num{0.0624}. This result substantially outperforms existing open-source baselines, including Qwen3.5-2B (\num{0.1532}), Qwen3-VL-2B (\num{0.1821}), and FireRed-OCR (\num{0.2505}).

When breaking down the performance by language family, we observe several key strengths:
\begin{itemize}[leftmargin=1.4em, itemsep=0.2em, topsep=0.3em]
\item \textbf{Latin-Alphabet Settings}: ABot-OCR attains near-saturation accuracy across multiple European and Southeast Asian languages. Specifically, it achieves near-perfect transcription for German (\num{0.0005}), French (\num{0.0006}), Spanish (\num{0.0020}), Portuguese (\num{0.0030}), and Vietnamese (\num{0.0034}).
\item \textbf{Complex and Low-Resource Scripts}: More importantly, our model significantly reduces recognition errors on historically challenging writing systems. For example, on Arabic text, which features cursive connectivity and a right-to-left reading order, ABot-OCR lowers the edit distance to \num{0.0180}, compared to \num{0.1823} for Qwen3-VL-2B and \num{0.3895} for FireRed-OCR. Similarly, it handles the intricate vocalization marks of Thai (\num{0.0100}) and the diverse character sets of Russian (\num{0.1925}) with high precision.
\item \textbf{East Asian Typography}: For CJK languages, ABot-OCR also establishes a new state-of-the-art among the compared systems, leading on both Japanese (\num{0.1731}) and Korean (\num{0.2206}). This success is particularly notable because East Asian documents usually present severe challenges due to their vertical layouts, dense character spacing, and high glyph complexity.
\end{itemize}

In conclusion, these comprehensive benchmarks demonstrate that ABot-OCR generalizes robustly across diverse writing systems. By effectively resolving the structural and visual ambiguities of non-Latin and mixed-script layouts, our model proves to be highly reliable for global-scale document parsing.

\begin{table}[htbp]
\centering
\caption{Multilingual performance comparison (Edit Distance)}
\label{tab:multilingual_transposed}
\small 
\sisetup{
    table-format=1.3,
    round-mode=places,
    round-precision=3,
    detect-weight = true,    
    detect-inline-weight = math, 
    table-number-alignment=center
}
\setlength{\tabcolsep}{2.8pt} 
\renewcommand{\arraystretch}{1.05}

\begin{tabular}{@{}l *{11}{S}@{}}
\toprule
\textbf{Model} &
\textbf{Arabic} &
\textbf{German} &
\textbf{Spanish} &
\textbf{French} &
\textbf{Japan} &
\textbf{Korean} &
\textbf{Portug} &
\textbf{Russian} &
\textbf{Thai} &
\textbf{Vietna} &
\textbf{Overall} \\
\midrule
Qwen3-VL-2B & 0.3895 & 0.0255 & 0.0326 & 0.0092 & 0.3140 & 0.4014 & 0.0336 & 0.3576 & 0.2299 & 0.0272 & 0.1821 \\
Qwen3.5-2B  & 0.1823 & 0.0154 & 0.0229 & 0.0370 & 0.3058 & 0.4010 & 0.0191 & 0.3432 & 0.1906 & 0.0210 & 0.1532 \\
FireRed-OCR & 0.3205 & 0.0184 & 0.0321 & 0.1206 & 0.3979 & 0.5258 & 0.0376 & 0.5227 & 0.4416 & 0.0877 & 0.2505 \\
\textbf{ABot-OCR (Ours)} & \bfseries 0.0180 & \bfseries 0.0005 & \bfseries 0.0020 & \bfseries 0.0006 & \bfseries 0.1731 & \bfseries 0.2206 & \bfseries 0.0030 & \bfseries 0.1925 & \bfseries 0.0100 & \bfseries 0.0034 & \bfseries 0.0624 \\
\bottomrule
\end{tabular}
\end{table}

\subsubsection{Ablation study on DHDO strategies}

In this section, we investigate how different DHDO configurations and training recipes affect end-to-end document parsing performance. Specifically, we study preference optimization signals across three primary document elements: text, tables, and formulas (\LaTeX). We also evaluate a generic GRPO baseline and a mixed configuration. Table~\ref{tab:gdpo_ablation} summarizes these findings.

Compared to the supervised baseline, the generic GRPO model improves the Overall score to 92.09 and lowers the TextEdit distance to 0.043. It also brings modest gains to table parsing. However, it slightly reduces the Formula\textsuperscript{CDM} score from 95.01 to 94.82. This dropdown indicates that a generic preference signal cannot uniformly strengthen brittle, highly structured outputs like mathematical equations.

Among all single-signal experiments, Table DHDO yields the largest improvements in table parsing and achieves the highest Overall score of 92.49. It secures a TEDS score of 86.95, a TEDS\textsubscript{s} score of 90.12, and a reading-order error of 0.142. Conversely, it provides minimal impact on text and formula metrics, proving its effect is highly specialized.

Text DHDO delivers the best text accuracy with a TextEdit score of 0.040. It also reaches the highest Formula\textsuperscript{CDM} score of 95.19 and the lowest single-signal reading-order error of 0.139. However, it leaves the table metrics nearly unchanged compared to the baseline model, showing a clear performance trade-off.

LaTeX DHDO successfully improves the TEDS\textsubscript{s} score to 89.14, demonstrating its utility in mathematical layouts. Unfortunately, it also increases the TextEdit error to 0.047. This negative trade-off suggests that over-concentrated formula preferences can interfere with plain-text accuracy if the training terms are not carefully balanced.

Finally, mixing all three optimization streams with a 1:1:1 ratio achieves the absolute best performance across the board. It delivers the highest Overall score of 93.30 and the strongest joint profile, including a TextEdit of 0.037, a TEDS of 88.83, a TEDS\textsubscript{s} of 91.94, a reading-order error of 0.133, and a Formula\textsuperscript{CDM} of 94.86. These comprehensive comparisons strongly support our final design choice. Therefore, we select task-decomposed preference learning with a mixed optimization schedule as our default training strategy for full-document parsing.

\begin{table}[htbp]
\centering
\caption{Ablation study on DHDO strategies.}
\label{tab:gdpo_ablation}
\begin{tabular}{l
    S[table-format=2.2]   
    S                     
    S                     
    S                     
    S                     
    S}                    
\toprule
\textbf{Methods} & 
\multicolumn{1}{c}{\textbf{Overall↑}} & 
\multicolumn{1}{c}{\textbf{TextEdit↓}} & 
\multicolumn{1}{c}{\textbf{Formula\textsuperscript{CDM}↑}} & 
\multicolumn{1}{c}{\textbf{Table\textsuperscript{TEDS}↑}} & 
\multicolumn{1}{c}{\textbf{Table\textsuperscript{TEDS\(_s\)}↑}} & 
\multicolumn{1}{c}{\textbf{R-order\textsuperscript{Edit}↓}} \\
\midrule
\multicolumn{7}{l}{\textit{Baselines}} \\
Base  & 91.71 & 0.049 & 95.01 & 85.04 & 88.14 & 0.153 \\
Base + GRPO  & 92.09 & 0.043 & 94.82 & 85.75 & 88.73 & 0.152 \\
\midrule
\addlinespace
Base + Table DHDO  & 92.49 & 0.045 & 95.03 & 86.95 & 90.12 & 0.142 \\
Base + Text DHDO  & 92.10 & 0.040 & 95.19 & 85.10 & 88.25 & 0.139 \\
Base + LaTeX DHDO & 92.03 & 0.047 & 94.95 & 85.83 & 89.14 & 0.147 \\
Base + Mix 1:1:1 & 93.30 & 0.037 & 94.86 & 88.83 & 91.94 & 0.133 \\
\bottomrule
\end{tabular}
\end{table}

\section{Conclusion}

In this work, we presented ABot-OCR, an end-to-end vision-language framework that treats document parsing as a direct image-to-Markdown generation task. We designed a robust three-stage training recipe. First, we perform modular document parsing to establish a strong foundational perception of glyphs, layouts, formulas, and tables. Second, we apply supervised specialization using normalized page-level Markdown data. Third, we implement Decoupled Heterogeneous Document Optimization (DHDO) during post-training. This final stage uses structure-constrained reinforcement learning to strictly enforce textual fidelity and markup well-formedness. Furthermore, we develop a dedicated data engine to supply the large-scale, structurally consistent training labels required by this entire pipeline.
Our extensive experiments validate the effectiveness of this design. On the OmniDocBench v1.5 and v1.6 benchmarks, ABot-OCR attains state-of-the-art scores of 92.81 and 93.30 among all end-to-end systems. In doing so, it substantially narrows the performance gap between end-to-end models and strong pipeline baselines. Additionally, our multilingual evaluations across ten diverse languages confirm that the framework generalizes robustly across different writing systems.
In future work, we plan to improve our model's inference efficiency. We will also extend our multilingual parsing capabilities to handle complex document layouts with even richer structural diversity.

\section*{Acknowledgments}
We would like to thank Mr. Chunlong Lv and his team from the POI division of the Amap Data Business Unit for providing real-world OCR data support for this work.

\bibliographystyle{plainnat}
\bibliography{main}

\newpage
\beginappendix

\section{Qualitative Examples}
The following cases demonstrate the model’s capability to bridge the gap between raw pixels and structured semantic understanding.

\subsubsection{perception capability}
\begin{figure}[H]
    \centering
    \begin{minipage}[b]{0.48\textwidth}
        \centering
        \includegraphics[width=\linewidth]{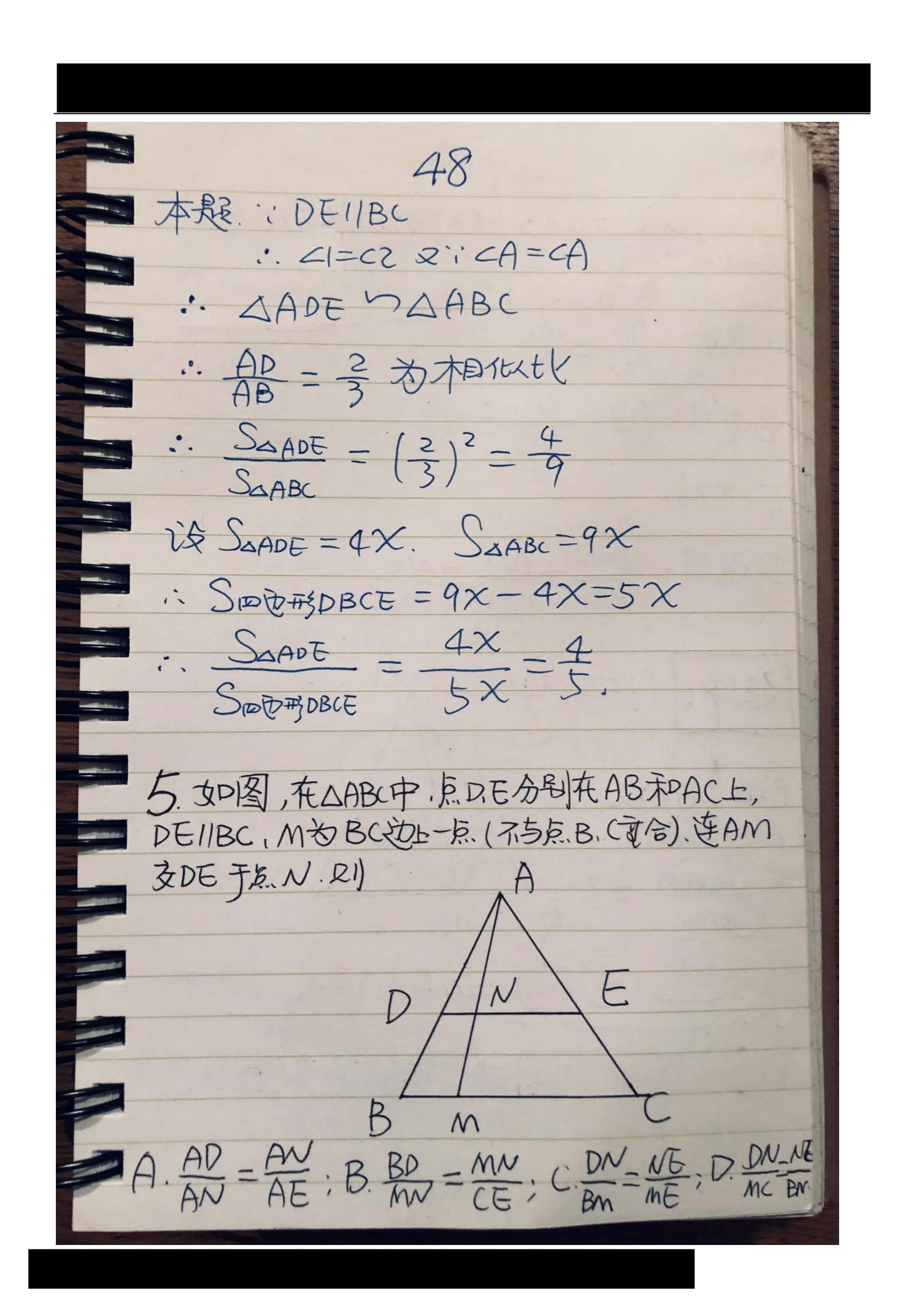}
        \caption{Input: original image}
        \label{fig:text_rec_left}
    \end{minipage}
    \hfill
    \begin{minipage}[b]{0.48\textwidth}
        \centering
        \includegraphics[width=\linewidth]{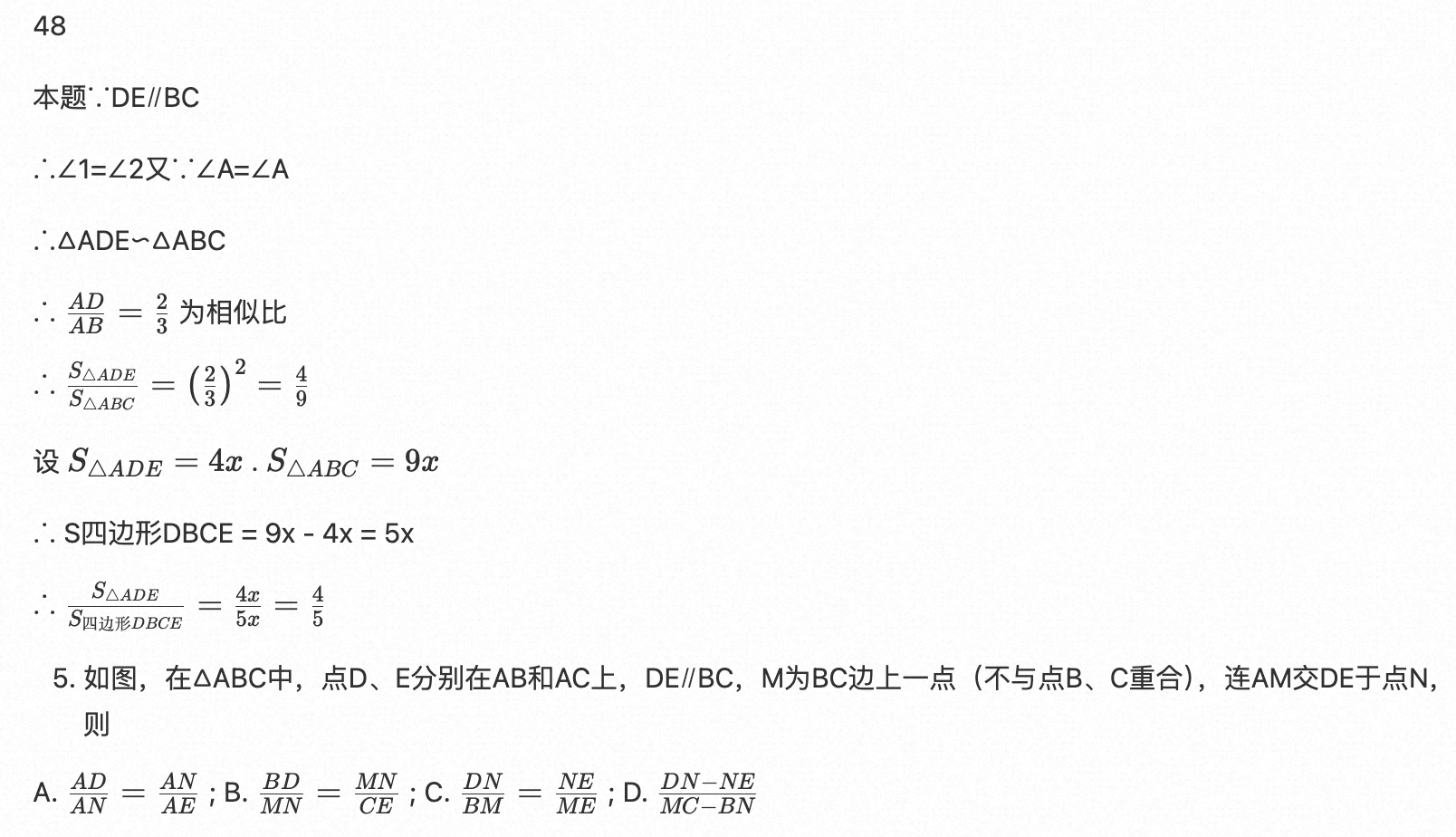}
        \caption{Output: Rendered \LaTeX{} Result}
        \label{fig:text_rec_right}
    \end{minipage}
\end{figure}

\subsubsection{structural capability}
\begin{figure}[H]
    \centering
    \begin{minipage}[b]{0.48\textwidth}
        \centering
        \includegraphics[width=\linewidth]{figure/formular_example.png}
        \caption{Input: original image}
        \label{fig:math_rec_left}
    \end{minipage}
    \hfill
    \begin{minipage}[b]{0.48\textwidth}
        \centering
        \includegraphics[width=\linewidth]{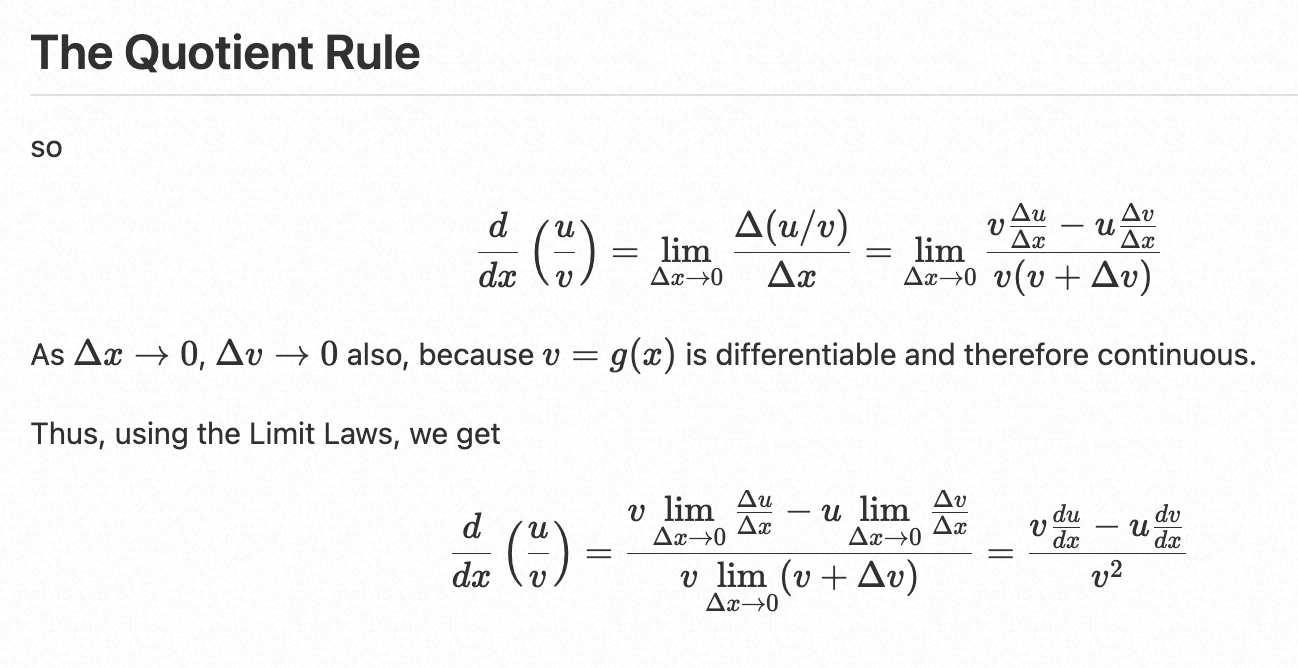}
        \caption{Output: Rendered \LaTeX{} Result}
        \label{fig:math_rec_right}
    \end{minipage}
\end{figure}

\begin{figure}[H]
    \centering
    \begin{minipage}[b]{0.48\textwidth}
        \centering
        \includegraphics[width=\linewidth]{figure/table.png}
        \caption{Input: original image}
        \label{fig:table_rec_left}
    \end{minipage}
    \hfill
    \begin{minipage}[b]{0.48\textwidth}
        \centering
        \includegraphics[width=\linewidth]{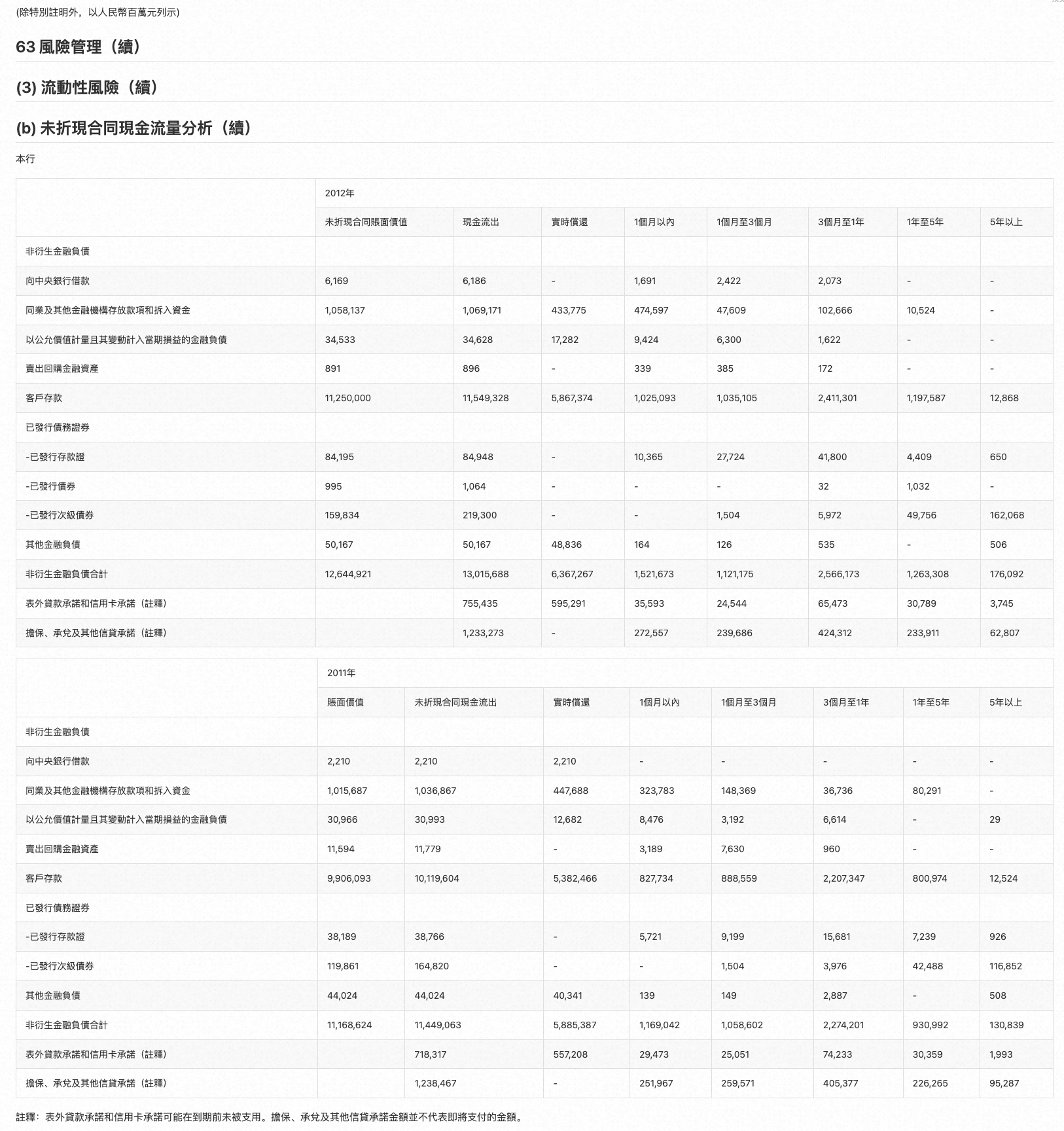}
        \caption{Output: Rendered \LaTeX{} Result}
        \label{fig:table_rec_right}
    \end{minipage}
\end{figure}

\begin{figure}[H]
    \centering
    \begin{minipage}[b]{0.48\textwidth}
        \centering
        \includegraphics[width=\linewidth]{figure/news_layout.jpg}
        \caption{Input: original image}
        \label{fig:doc_left}
    \end{minipage}
    \hfill
    \begin{minipage}[b]{0.48\textwidth}
        \centering
        \includegraphics[width=0.48\linewidth]{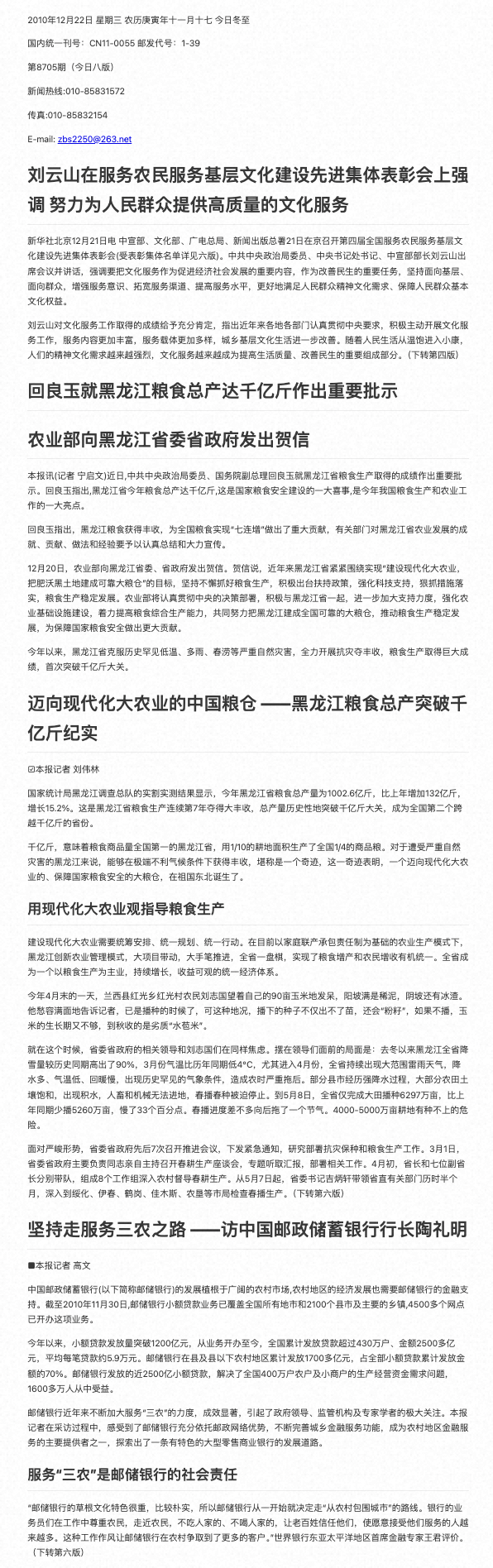}
        \hfill
        \includegraphics[width=0.48\linewidth]{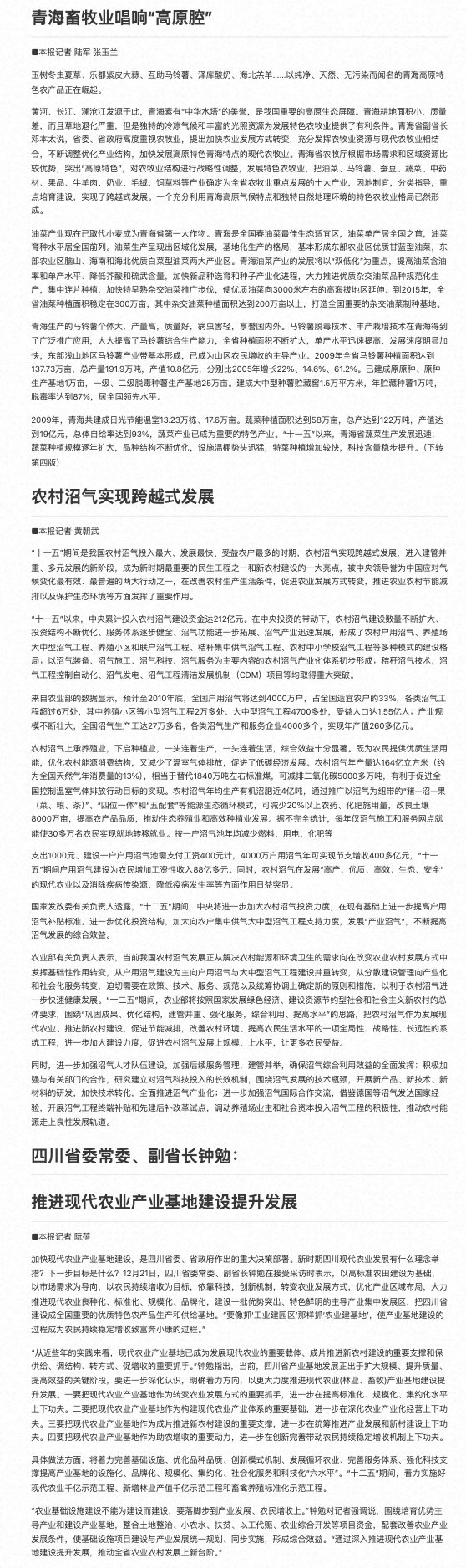}
        
        \caption{Output: Rendered \LaTeX{} Result}
        \label{fig:doc_right-pair}
    \end{minipage}
\end{figure}
Due to the excessive length of the document, we have divided the output into two parts for presentation.


\end{document}